**Addressing crash-imminent situations caused by human driven vehicle errors in a mixed traffic stream: a model-based reinforcement learning approach for CAV**

**Jiqian Dong**
Graduate Research Assistant, Center for Connected and Automated Transportation (CCAT), and Lyles School of Civil Engineering, Purdue University, West Lafayette, IN, 47907.
Email: dong282@purdue.edu
ORCID #: 0000-0002-2924-5728

**Sikai Chen\***
Visiting Assistant Professor, Center for Connected and Automated Transportation (CCAT), and Lyles School of Civil Engineering, Purdue University, West Lafayette, IN, 47907.
Email: chen1670@purdue.edu; and
Visiting Research Fellow, Robotics Institute, School of Computer Science, Carnegie Mellon University, Pittsburgh, PA, 15213.
Email: sikaichen@cmu.edu
(Corresponding author)
ORCID #: 0000-0002-5931-5619

**Samuel Labi**
Professor, Center for Connected and Automated Transportation (CCAT), and Lyles School of Civil Engineering, Purdue University, West Lafayette, IN, 47907.
Email: labi@purdue.edu
ORCID #: 0000-0001-9830-2071

## ABSTRACT

It is anticipated that the era of fully autonomous vehicle operations will be preceded by a lengthy "Transition Period" where the traffic stream will be mixed, that is, consisting of connected autonomous vehicles (CAVs), human-driven vehicles (HDVs) and connected human-driven vehicles (CHDVs). In recognition of the fact that public acceptance of CAVs will hinge on safety performance of automated driving systems, and that there will likely be safety challenges in the early part of the transition period, significant research efforts have been expended in the development of safety-conscious automated driving systems. Yet still, there appears to be a lacuna in the literature regarding the handling of the crash-imminent situations that are caused by errant human driven vehicles (HDVs) in the vicinity of the CAV during operations on the roadway. In this paper, we develop a simple model-based Reinforcement Learning (RL) based system that can be deployed in the CAV to generate trajectories that anticipate and avoid potential collisions caused by drivers of the HDVs. The model involves an end-to-end data-driven approach that contains a motion prediction model based on deep learning, and a fast trajectory planning algorithm based on model predictive control (MPC). The proposed system requires no prior knowledge or assumption about the physical environment including the vehicle dynamics, and therefore represents a general approach that can be deployed on any type of vehicle (e.g., truck, buse, motorcycle, etc.). The framework is trained and tested in the CARLA simulator with multiple collision imminent scenarios, and the results indicate the proposed model can avoid the collision at high successful rate (>85%) even in highly compact and dangerous situations.

**Keywords:** Connected Autonomous Vehicle, Crash avoidance, Model-based Reinforcement Learning

# INTRODUCTION

Traffic safety is still a global concern due to the high number of fatalities and severe injuries incurred from traffic accidents. According to 2018 statistics released by WHO (*1*), annual road traffic deaths has reached 1.35 million and the road traffic accidents is the leading cause of death of children and young adults from 9-25. Besides, the economic losses due to road traffic crashes cost most countries 3% of their gross domestic product (*2*). There is widespread optimism that the advent of Autonomous Vehicles (AVs) can help mitigate the problem of safety. AV technology has advanced rapidly in recent years with several fully autonomous models and a number of autonomous features already existing on the market. With their faster decision process, small reaction time, and higher accuracy in control, AVs are expected to enhance the traffic safety by eliminating participation of humans and corresponding errors. Therefore, automated driving systems are expected to directly reduce as much as 94% of all accidents (*3–5*). In addition, the connectivity feature of connected and autonomous vehicles (CAVs) will facilitate vehicle-to-vehicle (V2V) communication which can further enhance the performance of the automated driving systems by facilitating the dissemination of traffic-related information. Connectivity devices generally have longer range compared with on-board sensors and therefore can help the CAV undertake proactive actions to avoid imminent crashes. Meanwhile, V2V connectivity does not prone to occlusion or inclement weather and contains less noises. This longer, faster and higher-quality information could generally improve the decision process of AVs, especially in the adversarial situations (*6*). In summary, the AV with connectivity features, often referred as Connected Autonomous Vehicles (CAVs), are believed as the key to reach Vision Zero (zero accident) (*7*).

　　Technology forecasters hold the view that there is still a long way before achieving full automation with 100% market penetration. Therefore, subsequent to the introduction of AV on roadways, it is anticipated that there will be a prolonged phase (typically referred to the "transition period") where there will be mixed traffic (CAVs, human driven vehicles (HDVs), and connected human driven vehicles (CHDVs)). Even though CAVs are expected to lead to safer roads in the long run, in the short run, the mixed nature of technology and the growing pains of AV adoption and deployment are likely to cause an uptick in crashes before the safety benefits start to outpace the initial increase in crashes during the transition period. In the early part of the transition period, the inevitable driving errors exhibited by human drivers will potentially cause difficulties to AVs operations in the mixed stream, particularly where there are no AV-dedicated lanes. Currently, research on vehicle automation safety focus primarily on improving the onboard system so that the AV can operate without compromising the safety of the neighboring HDVs (*8–10*) but have little guidance on the opposite direction: helping the AV operate without its safety being compromised by the neighboring HDVs. In other words, the existing literature does not address the situations where other HDVs are at fault and may cause collisions with the ego-AV (the AV in question). In this research, we specifically address the development of an AV control system that can be installed in the AV to mitigate safety-critical situations caused by the errant HDVs in the neighborhood of the AV.

　　Another significant issue regarding AV system safety is the lack of safety-critical data (*11*). Although many commercial autonomous driving companies have claimed success in autonomous driving experiments by showing the increasing time during which the vehicle was self-driven, their solution for improving the AV models is brute force in nature: feeding the model with more real-world driving data. However, since the data used for training is primarily collected using human driver subjects, the generated datasets are sparse regarding "extreme cases" or "safety critical" situations that could potentially lead to crashes. An apparent dilemma is that even the existing AV models have been fully tested successfully in the normal driving situations, and the question of whether they can handle those extreme cases remains unanswered. A natural way to solve the problem is to use simulated driving environments which are often realistic enough to describe the natural driving environment (NDE) and generate those imminent extreme-case situations. Recently, with the development of urban driving simulators such as CARLA(*12*) and SUMMIT(*13*), the cost for generating collision pertained trajectories and scenarios have reduced significantly. This offers great promise for collecting simulation data and for developing data-driven solutions.

**Model based reinforcement learning**

Typically, in robotics research, crash avoidance is often defined as a trajectory planning problem, which requires the robots (vehicles) to make sequential decisions to navigate through the operational environment. Two successful approaches to solve this problem are planning and reinforcement learning (RL) (*14*). The former uses "known" state transition model and dynamic programming methods for generating optimal control commands (*15*), while the latter "learns" to approximate a global value or policy function (*16*) and use the value or policy function to generate decisions. Planning method is generally formulated as an optimization-based method that requires no data or learning process, but the state transition model (system dynamics) need to be explicitly defined. For classic planning methods in autonomous driving, these state transition models are built on top of handcrafted features and models describing the physical environment (*17, 18*), which require a full understanding of the kinematic properties of all surrounding vehicles. However, the vehicle dynamics are generally difficult enumerate since there typically exist a large number of vehicles with different features such as shape, weight, tire condition, etc., and all these features could affect the dynamic model. Therefore, it is not prudent to use a single physical model to describe all the surrounding vehicles which could potentially restrict the use cases in the real-world deployment. Also, another shortcoming for such physical model-based methods is the non-consideration of vehicle-to-vehicle interactions, which can be alleviated using learning based methods (*19, 20*).

On the other hand, the RL method, does not necessarily need to know the state transition model, but requires collecting a massive dataset for training the agent (*14*). Even though the simulation environment can mitigate the problem of data scarcity, it is still impossible to enumerate all the potential cases that the AV can encounter, and therefore reliability of the RL model is still hard to guarantee. Also, majority of RL algorithms are combined with black box deep neural networks, which can exacerbate the problem of user trust especially under imminent situations. The intersection of two methods yields the "model-based reinforcement learning" (model based RL), which leverage the data to first estimate the state transition model and then conducting planning based on the estimated model. This combined method reaps the benefits from both methods: data/training efficient and model agnostic, which motivates us exploring such method in this research.

In general, the model based RL method for AV trajectory planning contains 2 modules: state prediction and path planning. State prediction performs as an estimation to the physical environment, which specifically addresses the problem of reasoning the future states based on the previous information. In other words, it "tells" what the state (locations, speed, acceleration, etc.,) the surrounding objects will reach in a short future (prediction horizon) based on the historical trajectory. It is critical since it is the first step in the entire trajectory planning task and the prediction error can generally propagates or even amplified in the subsequent path planning. In this work, we utilize the deep learning based method to conduct state prediction.

With regard to the path planning module, it is built on top of the state prediction model. Since the state prediction cannot be perfect, the ideal planning module should stop the error propagation by robustly outputting a safe path even with the inaccurate state prediction. Secondly, it should be adaptive to the highly dynamic scenarios, especially under the cases when new agents emerge (a pedestrian suddenly crosses the road or aggressive lane change of surrounding vehicles). Model Predictive Control (MPC) is a common control approach is a general approach that meets these 2 criteria for generating collision free paths (*21–24*). The key idea is to "re-plan" at each time step and only execute the first step of the current optimal trajectory. Since the feasibility of actions is evaluated at each time step, this method is capable of handling the rapidly changing scenarios. Classic MPC in control theory seeks to formulate the planning problem into a complicated optimization problem with "given" physical plant model (system dynamics). In our model-based RL setting, we apply MPC with the data-driven state prediction module and utilize a fast and simple plan algorithm to replace the heavy optimization. Overall, the benefits of the proposed method are: data efficient, model explainable, stable, transferrable across scenarios (robust).

**Main contribution**
In summary, the main contributions of study are:
- Create and an Open AI gym (*25*) interface for reinforcement learning in CARLA simulator.
- Develop a fast model-based reinforcement learning method to control the CAV under crash imminent situations.
- Build an end-to-end flow containing data collection, training, testing, and retraining.
- Conduct multiple experiments for different imminent adversarial scenarios and test the robustness of the model.

After conducting a large number of experiments, it was observed that the proposed workflow works generally well for many unseen scenarios and can be easily adapted to scenarios with different agents.

**METHODS**
As introduced before, the entire methods contain 2 main modules, state prediction and path planning. The former seeks to learn a map from previous and current states to the future state, which performs as a physical engine to describe the driving environment. The latter utilizes state prediction model to rehearse the planned trajectory and pick the feasible one to avoid the crash.

**State prediction module**
Under adversarial driving situations, the CAV needs to make instantaneous decisions within a short time to avoid collisions. This requires the model to compute fast, and thus precludes using heavy neural networks. In this work, 3 network architectures are experimented: 3-layer fully connected neural network (FCN), single layer long short-term memory network (LSTM), single layer FCN (linear regression).

At each timestep, each state prediction model takes a sequence of historical states $X_t \in \mathcal{S}$ which contains previous $n$ timesteps' trajectory of that agent, and predict a new state $\hat{X}_{t+1} \in \mathcal{S}$ for timestep $t + 1$, where $\mathcal{S}$ is the state space.

$$\hat{X}_{t+1} = f(X_t) \in \mathcal{S} \qquad (1)$$

With regard to the prediction model $f$, 3 network architectures are experimented in this research: 3-layer fully connected neural network (FCN), single layer long short-term memory network (LSTM), single layer FCN (linear regression). It may be noted that the prediction model for the ego CAV and surrounding vehicles are different since the historical control inputs for CAV are known and can be leveraged for more accurate prediction. Therefore, for CAV, $\hat{X}_{t+1} = f_{CAV}(X_t, A_t)$; for HDVs, $\hat{X}_{t+1} = f_{HDV}(X_t)$.

Since the number of vehicles in the proximity of CAV keeps changing, using one single (centralized) prediction model to predict the motion of all the surrounding vehicles will require the model to handle dynamic lengths of inputs. This will inevitably complicate the model and requires deeper neural networks, which may conflict the requirement of low computation cost. The simplest way to solve the problem is using decentralized manner which associates each agent with a separate state prediction model.

The loss function for training the networks is the Mean Squared Error (MSE) loss, which is defined in **Equation 2**. $b$ represents the batch size.

$$\mathcal{L} = \frac{1}{b}\sum_{i=1}^{b}\left\|\hat{X}^i_{t+1} - X^i_{t+1}\right\|^2 \quad (2)$$

**Path planning module**
In general, the path planning module is formulated as an optimization problem with an objective (cost) function and constraints. The objective function reflects the value of the current state and current and future actions. For the collision imminent situations, the primal goal for the CAV is to quickly compute a collision free path. Therefore, in this work, we apply the classic safety indicator: time to collision (TTC) as the cost function. Using the simplist circle algorithm (26), the computation logic of a 2D TTC between a pair of vehicles $(i, j)$: $TTC_{ij}$ is as follows:

$$\left[x_i + v_{xi} \cdot TTC_{ij} + \frac{1}{2}a_{xi} \cdot TTC_{ij}^2 - (x_i + v_{xj} \cdot TTC_{ij} + \frac{1}{2}a_{xj} \cdot TTC_{ij}^2)\right]^2 + \left[y_i + v_{yi} \cdot TTC_{ij} + \frac{1}{2}a_{yi} \cdot TTC_{ij}^2 - (y_i + v_{yj} \cdot TTC_{ij} + \frac{1}{2}a_{yj} \cdot TTC_{ij}^2)\right]^2 = \left(R_i + R_j + d_s\right)^2 \quad (3)$$

Where $x_i, v_{xi}, a_{xi}$ represent the location, speed and acceleration for $i^{th}$ agent along the $x$ direction, respectively; similar notations are applied for $j^{th}$ vehicle and along the $y$ direction; $R_i$ and $R_j$ stand for the radius of the circle standing for each vehicle; $d_s$ is minimum the safety distance. This vanilla definition contains unknowns on both sides, and because of the distance computation, getting $TTC_{ij}$ requires solving a quadratic equation. Meanwhile, since the planning method is an MPC-based approach (the TTC needs to be computed with the new states at each timestep), some extent of error is tolerable. Also in reality, obtaining accurate acceleration of vehicles is technically hard. We therefore simplify the scenario by assuming a constant speed (0 acceleration) for each vehicle in TTC computation. After reorganizing, the equation (2) can be simplified. After using the vector notation, the closed form for TTC is:

$$TTC_{ij} = -\frac{\|\overrightarrow{\Delta x_{ij}}\| - d_s}{proj(\overrightarrow{\Delta v_{ij}}, \overrightarrow{\Delta x_{ij}})} \quad (4)$$

Where the $proj(\vec{a}, \vec{b}) = \frac{\vec{a} \cdot \vec{b}}{\|\vec{b}\|}$ computes the projection of vector $\vec{a}$ on the direction of $\vec{b}$, $\overrightarrow{\Delta x_{ij}}$, $\overrightarrow{\Delta v_{ij}}$ represent the location difference and velocity difference between vehicle $i$ and vehicle $j$ in a vector form. In this work since we only consider the TTC between the ego-CAV and surrounding HDVs, $i$ represents the ego-CAV. We fold the vehicle shape circle $R_i$ and $R_j$ into the safety distance $d_s$ and use a large value for this variable to add more safety redundancy.

Similarly, to guarantee collision-free conditions from static objects, we also add the TTC computation between the ego-CAV to all the surrounding obstacles including vehicles and static object $TTC_{is}$ existing on the roadsides.

$$TTC_{is} = -\frac{\|\overrightarrow{\Delta x_{is}}\| - d_s}{proj(\overrightarrow{v_i}, \overrightarrow{\Delta x_{is}})} \quad (5)$$

The only difference is the projection in denominator, for the static objects, only the CAV's velocity ($\overrightarrow{v_i}$) needs to be considered. The raw TTC values (both $TTC_{ij}$ and $TTC_{is}$) represent the time before collision between ego-CAV and all the surrounding objects. We need to specifically focus on the small TTC values since they are more safety critical, therefore we consider a threshold $T_{safe}$ and set the TTC values above this value to be $+\infty$. Then the threshold TTC values are aggregated into a single value representing the cost (risk) of current state as shown in **Equation (6)**

$$cost = \left(\sum_i \frac{1}{TTC_{ij}}\right) + \lambda \left(\sum_s \frac{1}{TTC_{is}}\right) \quad (6)$$

**MPC for path planning**

For the general MPC method, the optimal control commands are computed by solving an optimization problem. However, in this settings, TTC cost is non-convex, and even if it is differentiable, the gradient must pass through the state prediction model to rectify the control commands, meaning the imperfection of the model can be amplified and cause severe flaws to the path planning. Therefore, here, we utilize the easiest gradient free method: random test and pick the elites. The logic of this method is rather straightforward and contains the following 4 steps: (1) At each time step, generate $n$ sequence with each sequence contains $h$ actions, where $n$ is the number of tested trajectories and $h$ is the planning horizon. (2) For each trajectory, sequentially feed the total $h$ actions into the states prediction model to compute the future states and costs at each step. (3) Aggregate the costs for each trajectory. (4) Pick the trajectory with lowest cumulative cost and execute the first action of that trajectory. Even if this is a simple method without computing gradient or using any advanced optimization techniques, the results proved that the CAV is able to figure out a sequence of actions that can successfully avoid the collision caused by aggressive HDVs. The advantages of such method are the high flexibility and robustness. It is able to be deployed in any of the traffic situations with different number of agents, different type of agents (sedan, van, buses, etc.,) and is robust across all the scenarios. Furthermore, the computation complexity only grows linearly with number of surrounding HDVs, which is not achievable for majority of optimization-based methods.

**Overall end-to-end algorithm**

The overall end-to-end algorithm contains 3 main phases: Warming up phase (for collecting data); Training phase (estimate the state prediction model) and Path planning phase (testing for crash avoidance). The detailed algorithm is documented in **Algorithm 1**. These 3 steps follow the process of standard model-based RL approaches: collecting experience, estimating the model, and planning with the estimated model. Noticing the experiences in the testing (planning) phase can be added back to the replay memory for retraining the model. This guarantees that the model can be improved even after deployment.

---

**Algorithm 1    Model based reinforcement learning method**

---

Initialize the replay memories $R_{HDV}$ and $R_{CAV}$ to store the transition for HDVs and CAV, respectively

Initialize the weights for state prediction network for both CAV ($f_{CAV}$) and HDVs ($f_{HDV}$)

*#1 Warming up steps (Data collection)*

For data collection timestep $t = 1$ to $T$ (warming up steps) **do**

        Take random action combination for CAV $A_t$

        Gather the state-action transition $(X_t, A_t, X_{t+1})$ for CAV

        Gather the state transition $(X_t, X_{t+1})$ for HDVs

        Store the transitions $(X_t, X_{t+1})$ into the memory buffer $R$

---

*#2 Training loop (Estimate the state prediction model)*

For training timestep $t$ to $T$ (training steps) **do**

        Read HDV transitions batches $\left\{X_t^i, X_{t+1}^i\right\}_{i=1}^b$ from the memory buffer $R_{HDV}$

        Compute the loss for HDV state prediction model: $\mathcal{L}_{HDV} = \frac{1}{b}\sum_{i=1}^b \left\| f_{HDV}(X_t^i) - {X^i}_{t+1} \right\|^2$

        Backpropagate $\mathcal{L}_{HDV}$ and update the weight for $f_{HDV}$

        Read CAV transitions batches $\left\{X_t^i, A_t^i, X_{t+1}^i\right\}_{i=1}^b$ from the memory buffer $R_{CAV}$

        Compute the loss for CAV state prediction model: $\mathcal{L}_{CAV} = \frac{1}{b}\sum_{i=1}^b \left\| f_{CAV}(X_t^i, A_t^i) - {X^i}_{t+1} \right\|^2$

        Backpropagate $\mathcal{L}_{CAV}$ and update the weight for $f_{CAV}$

---

*#3 MPC Path planning (Crash avoidance)*

For executive horizon $t = 1$ to $T$ **do**

        Initial a cumulative cost array $\boldsymbol{J}$ for all $n$ trajectories

        For planning horizon $h = 1$ to $H$ **do**

                Random select $n$ trajectories with $n$ actions $\{A_h^i\}_{i=1}^n$

For each action $A_h^i$ in $n$ trajectories **do**

    Predict the future states for ego CAV: $\hat{X}^{CAV}{}_{h+1} = f_{CAV}(X_h, A_h^i)$

    Predict the future states all surrounding HDVs: $\hat{X}^{HDV}{}_{h+1} = f_{HDV}(X_h)$

    Compute the TTC based cost for each trajectory using **Equation (6)**

    Update the cumulative cost for each trajectory and saved in $J$

Pick the trajectories with lowest cumulative cost: $i^* = argmin(J)$

Execute the first action of $A_1^{i^*}$

Add $(X_t^{CAV}, A_1^{i^*}, X_{t+1}^{CAV})$ into $R_{CAV}, (X_t^{HDV}, X_{t+1}^{HDV})$

## EXPERIMENT SETTINGS
### Scenario settings
Among many crash imminent situations caused by the human drivers faults, the 2 cases (rear end and side impact) shown in **Figure 1** are most frequent (*27*) . These 2 situations are mainly originated from the illegal or aggressive lane changings from the grey-colored vehicles in the figure. This can happen in the real world when the red-colored vehicle is in the blind spot of the grey-colored vehicle.

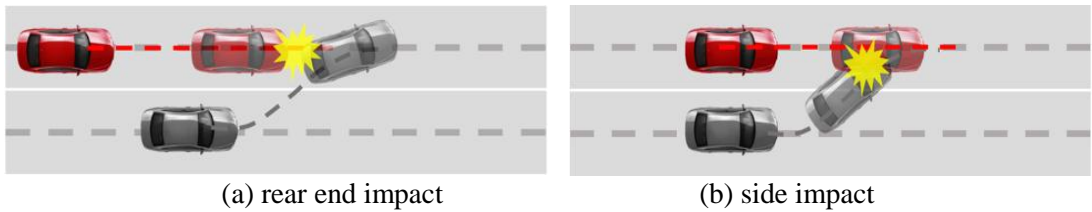

        (a) rear end impact                (b) side impact

**Figure 1. collision patterns (red: CAV, grey: HDV)**

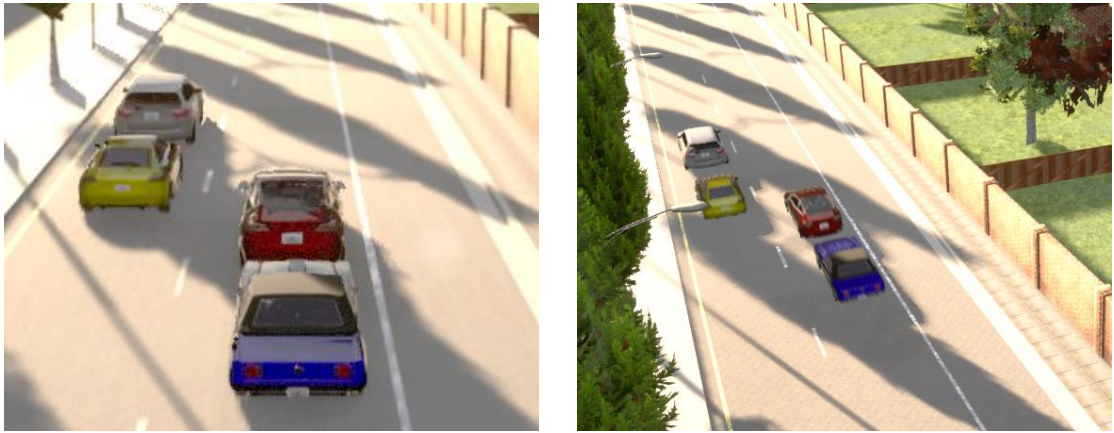

**Figure 2. Simulated situation in CARLA**

    To establish this most-frequent safety-imminent case in the CARLA simulator, we introduce 4 vehicles as shown in the **Figure 2**, the yellow vehicle represents the "at fault" HDV and the CAV is in red. This setting can be described as follows: the yellow HDV wish to overtake the grey vehicle but fail to identify the red vehicle (CAV) in its blind spot. This aggressive lane changing will likely to cause a crash particularly where the driving environment is compact (the CAV cannot brake hard since this will cause the rear end collision with the blue HDV). This would require the CAV to generate a sequence of maneuver to avoid the crash under this imminent situation. Note that in the simulation, other than the

cases shown in **Figure 2**, we also establish scenarios for yellow HDV to overtake from left and may cause a side-swipe collision with CAV on the right. This extra setting is to test the robustness of the model. The simulation step size is set as 0.05s/step (or 20 steps/s), and the aggressive overtaking maneuver for yellow HDV is generated manually by driving the vehicle with Logitech G27 Racing Wheel. To facilitate the RL experiments, we further develop a Open AI gym (*25*) interface with the Python API to connect CARLA simulator.

**State space**

The state describes the driving environment for both CAV and surrounding HDVs is the input fed into the trajectory prediction model to get the future states. For the state space in this research, we consider the historical location, speed, acceleration in both $x$ and $y$ directions for ego-CAV and surrounding vehicles in the past $n$ timesteps. Moreover, for ego-CAV, since the control commands are available, it should be included into the state for better state prediction. The state for HDVs and CAV are as follows:

$$X_{HDV} = \begin{bmatrix} x_1 & y_1 & v_{x1} & v_{y1} & a_{x1} & a_{y1} \\ \vdots & \vdots & \vdots & \vdots & \vdots & \vdots \\ x_n & y_n & ... & ... & a_{xn} & a_{yn} \end{bmatrix}_{n \times 6} \quad (6)$$

$$[X_{CAV}, A_{CAV}] = \begin{bmatrix} x_1 & y_1 & v_{x1} & v_{y1} & a_{x1} & a_{y1} & throttle_1 & steering_1 & break_1 \\ \vdots & \vdots & \vdots & \vdots & \vdots & \vdots & \vdots & \vdots & \vdots \\ x_n & y_n & ... & ... & a_{xn} & a_{yn} & throttle_n & steering_n & break_n \end{bmatrix}_{n \times 6} \quad (7)$$

In this work, we set the history window size $n = 5$.

**Action space**

As the control commands (throttle, brake pedal position and steering angle) are all continuous variables that can take infinite number of values, it is difficult to efficiently search such infinite action space. Also, some control commands should be constrained. For example, the steering angle should be constrained by "inertia", meaning the consecutive commands should be close since it is unsafe to make drastic changes within a short period of time. Also, the brake and gas pedal should be exclusive and cannot be pressed simultaneously. To reduce the action space and consider these interval constraints, we discretize the action space to make them discrete variable and use a fixed transfer logic to map the discretized variable back to original control commands. Specifically, we define 2 categorical variables for longitudinal control and latitudinal control, respectively. Each variable can take value from set $\{0, 1, 2\}$, The first variable determines whether current action should be gas, brake, or nothing (keep). For the steering angle, the categorical variable determines the incremental direction (left, right, or keep) of the steer wheel. Therefore, if the latitudinal control command is not zero, the steering will either subtract (turn left) 0.1 unit (this unit is defined by the simulator, the total range of the steering angle is -1 to 1 while -1 is the full left and 1 the full right) or add 0.1 unit (turn right) based on the current steering position. Under this setting, we shrink the overall action space to $3 \times 3 = 9$ actions, which can be easily searched.

$$longitudinal = \begin{cases} 0 & brake \\ 1 & keep \\ 2 & gas \end{cases} \quad (8)$$

$$latitudinal = \begin{cases} 0 & left \\ 1 & keep \\ 2 & right \end{cases} \quad (9)$$

**RESULTS**

**State prediction model**

**Figure 3** shows the training curves of all the three state prediction models for both CAV and HDVs. The loss is the mean squared error between the predicted state and ground truth state captured from actual trajectories. The models converge very fast after several epochs while the linear regression (linreg) model has the best prediction performance and convergence rate. The result indicates even with the easiest model linear regression, the prediction accuracy can be guaranteed, and it can even outperform other neural networks. The reason is that for the imminent situations in **Figure 2**, The timestep size is small (0.05s/step), thus all the vehicles have limited maneuvers that can be easily captured with a simple model. It worth noting that linear regression model converges so fast that can even be trained online after collecting trajectories for surrounding vehicles in the real-world deployment phase. This property facilitates the high robustness for the algorithm since it can always "learn" right away when facing new situations.

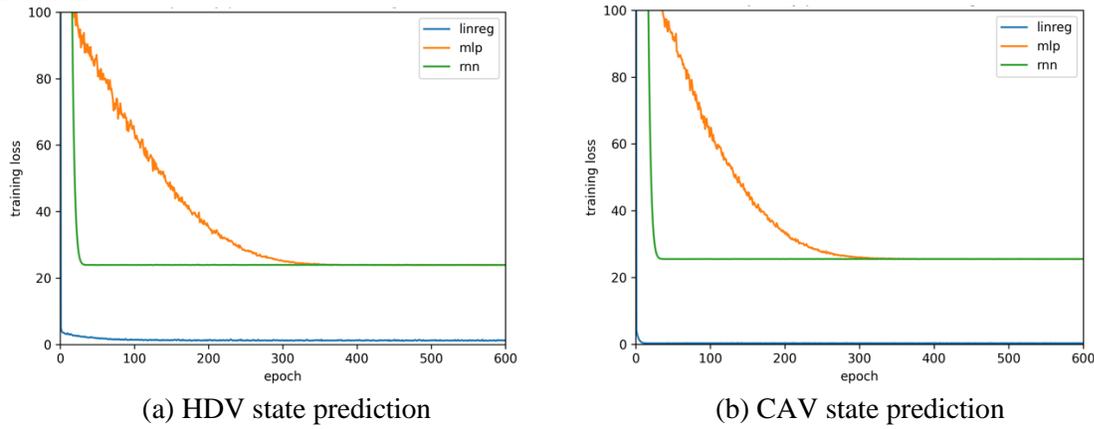

(a) HDV state prediction        (b) CAV state prediction

**Figure 3. Training curves for both state prediction model**

**Figure 4** shows the example of location prediction for both CAV and HDV in 5 random trajectories for the single-layered neural network (linear regression). The dash line in the figure represents the input trajectory (location in the past $n$ steps), the dot represents the location of the ground truth (label) and the star location is the predicted location. It is obvious that the CAV state prediction model has superior performance because the stars and dots have more coinciding cases. Also, the errors in the location prediction is comparatively low (lower than 1 meter), which we can be handled by the MPC based planning method.

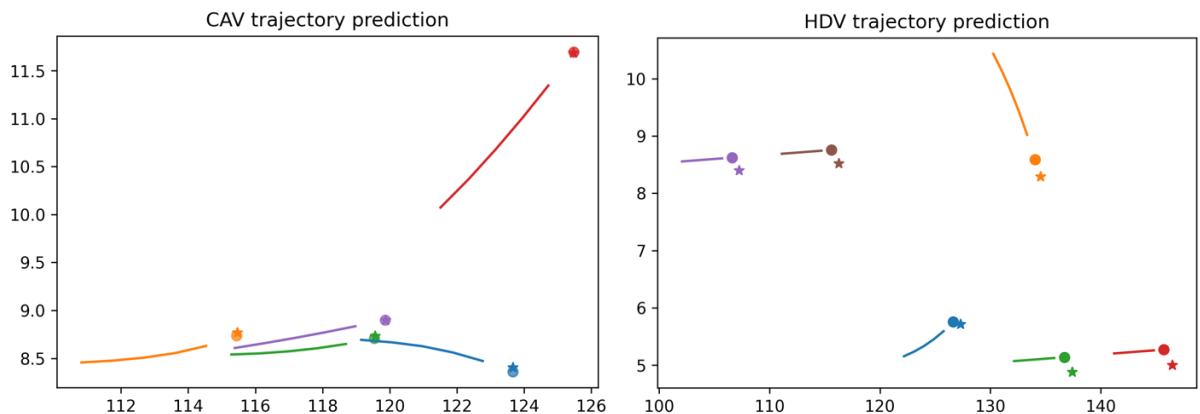

**Figure 4. Example location prediction in both x and y directions (Linear regression model)**

**Success rate**

The test cases we created are non-trivial so that there will be 100% crash if no proper evasive decisions maneuvers are executed. We adopt, as the main evaluation metric, the success rate for crash avoidance after CAV executes all the planned actions and fully stops. Since the linear regression model has the best performance in the state prediction, we test the proposed crash avoidance system built with this model under several multiple critical scenarios with different initial average speed: 15m/s, 20m/s, 25m/s; and different combinations of model parameters: number of random trajectories $n = 5, 10, 20, 30$ vs. planning horizons $h = 1, 3, 5, 7, 10$. Here, the average speed is the mean initial speed of all 4 vehicles, for each testing round, a random normal noise with zero mean and a standard deviation of $1 m/s$ is added to each vehicle. This is to guarantee the heterogeneity of the driving environment. Generally, the scenarios with higher initial average speed are riskier and have higher chance of collision. Each setting is tested 20 times, the average success rates of avoiding the crash vs. the model parameters are shown in **Figure 5**.

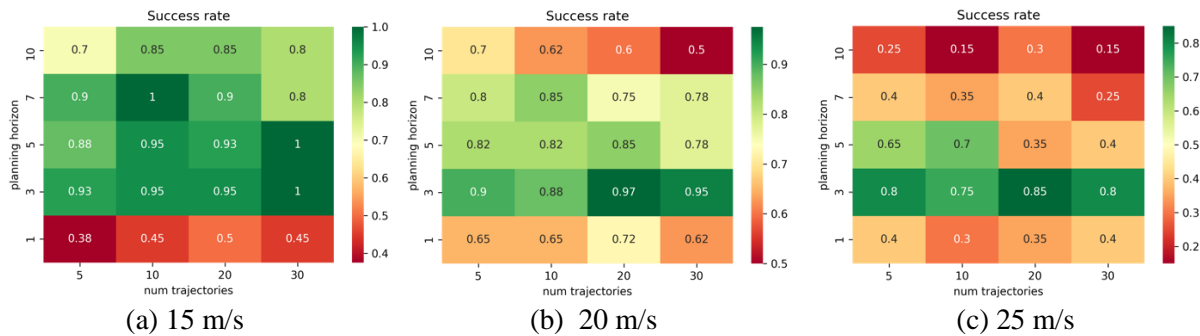

**Figure 5. Success rate for avoiding the crashes with different initial average speed**

From the **Figure 5**, it is obvious that the model parameters (number of trajectories and planning horizon) can greatly affect the avoidance rate. An increase in the number of trajectories can generally enhance the performance because more trials with more future possibilities can be evaluated. However, more trajectories means more computation is needed. With regard to the planning horizons, from the results above, it can be observed that either too large or too small planning horizon can harm the model performance in terms of the success rate. The optimal planning horizon is 3 steps (0.15s). The reason is that all the scenarios belong to crash-imminent situations, planning long into the future can be of no use because the crash will happen within a short period. Also, since the prediction error caused by discrepancies between the predicted trajectories and the actual trajectories driven by both HDVs and CAV is inevitable, longer planning horizon can induce larger drift. Furthermore, after testing multiple rounds, the experiences can be saved to further train the state prediction model. This property facilitates the improvement of the framework even after deployment.

**CONCLUDING REMARKS**

In this research, we developed a simple, fast model-based reinforcement learning framework for CAV to avoid collision caused by the external human error. The framework contains 2 cooperative submodules (state prediction and path planning) that can be trained end-to-end and has the capability of being improved online after deployment. The framework is a data-driven approach that does not require any prior knowledge or assumptions on the vehicle dynamics or physical models, but also has benefits of data efficiency from conventional planning-based method. Therefore, the flexibility and robustness can be expected under various of safety imminent situations with different vehicle number and types (i.e., truck or bus). After being deployed and tested in the CARLA environment, the results demonstrate the proposed algorithm can reach a high success rate of crash avoidance at least 85% even in a highly compact driving environments which have 100% crash rate if without proper evasive maneuver.

Moving forward, the framework can be tested by incorporating more testing scenarios containing different crash patterns other than side-impact. Also with vehicle connectivity (*28–30*), cooperative maneuvers can be expected to have higher performance particularly in the compact and complicated driving environments (*31–35*). In other words, the CAV can take over the control for surrounding CHDVs to make collaborative decisions to avoid a crash. Therefore, the cooperative maneuver with more than 1 agent and large action space can be further investigated in future research.

## ACKNOWLEDGMENTS
This work was supported by Purdue University's Center for Connected and Automated Transportation (CCAT), a part of the larger CCAT consortium, a USDOT Region 5 University Transportation Center funded by the U.S. Department of Transportation, Award #69A3551747105. The map data is from OpenStreetMap contributors and is available from https://www.openstreetmap.org. The contents of this paper reflect the views of the authors, who are responsible for the facts and the accuracy of the data presented herein, and do not necessarily reflect the official views or policies of the sponsoring organization. This manuscript is herein submitted for PRESENTATION ONLY at the 2022 Annual Meeting of the Transportation Research Board.

## AUTHOR CONTRIBUTIONS
The authors confirm contribution to the paper as follows: all authors contributed to all sections. All authors reviewed the results and approved the final version of the manuscript.

**REFERENCES**

1. World Health Organization. *The Global Status Report on Road Safety 2018*. 2018.
2. Lagarde, E. Road Traffic Injuries. In *Encyclopedia of Environmental Health*.
3. Rahman, M. S., M. Abdel-Aty, J. Lee, and M. H. Rahman. Safety Benefits of Arterials' Crash Risk under Connected and Automated Vehicles. *Transportation Research Part C: Emerging Technologies*, 2019. https://doi.org/10.1016/j.trc.2019.01.029.
4. Noy, I. Y., D. Shinar, and W. J. Horrey. Automated Driving: Safety Blind Spots. *Safety Science*.
5. Ye, L., and T. Yamamoto. Evaluating the Impact of Connected and Autonomous Vehicles on Traffic Safety. *Physica A: Statistical Mechanics and its Applications*, 2019. https://doi.org/10.1016/j.physa.2019.04.245.
6. Ha, P., S. Chen, R. Du, J. Dong, Y. Li, and S. Labi. Vehicle Connectivity and Automation: A Sibling Relationship. *Frontiers in Built Environment*.
7. Zwetsloot, G. I. J. M., M. Aaltonen, J. L. Wybo, J. Saari, P. Kines, and R. Op De Beeck. The Case for Research into the Zero Accident Vision. *Safety Science*.
8. Kalra, N., and S. M. Paddock. Driving to Safety: How Many Miles of Driving Would It Take to Demonstrate Autonomous Vehicle Reliability? *Transportation Research Part A: Policy and Practice*, 2016. https://doi.org/10.1016/j.tra.2016.09.010.
9. Koopman, P., and M. Wagner. Autonomous Vehicle Safety: An Interdisciplinary Challenge. *IEEE Intelligent Transportation Systems Magazine*, 2017. https://doi.org/10.1109/MITS.2016.2583491.
10. Naranjo, J. E., C. González, R. García, and T. De Pedro. Lane-Change Fuzzy Control in Autonomous Vehicles for the Overtaking Maneuver. *IEEE Transactions on Intelligent Transportation Systems*, 2008. https://doi.org/10.1109/TITS.2008.922880.
11. Feng, S., X. Yan, H. Sun, Y. Feng, and H. X. Liu. Intelligent Driving Intelligence Test for Autonomous Vehicles with Naturalistic and Adversarial Environment. *Nature Communications*, 2021. https://doi.org/10.1038/s41467-021-21007-8.
12. Dosovitskiy, A., G. Ros, F. Codevilla, A. López, and V. Koltun. CARLA: An Open Urban Driving Simulator. *Proceedings of the 1st Annual Conference on Robot Learning*.
13. Cai, P., Y. Lee, Y. Luo, and D. Hsu. SUMMIT: A Simulator for Urban Driving in Massive Mixed Traffic. 2020.
14. Moerland, T. M., J. Broekens, and C. M. Jonker. Model-Based Reinforcement Learning: A Survey. *Proceedings of the International Conference on Electronic Business (ICEB)*, 2020.
15. Bertsekas, D. P. *Dynamic Programming and Optimal Control*. 2012.
16. Sutton, R. S., and A. G. Barto. Reinforcement Learning : An Introduction 2nd (19 June, 2017). *Neural Networks IEEE Transactions on*, 2017.
17. Houenou, A., P. Bonnifait, V. Cherfaoui, and W. Yao. Vehicle Trajectory Prediction Based on Motion Model and Maneuver Recognition. 2013.
18. Lefèvre, S., D. Vasquez, and C. Laugier. A Survey on Motion Prediction and Risk Assessment for Intelligent Vehicles. *ROBOMECH Journal*.
19. Deo, N., and M. M. Trivedi. Convolutional Social Pooling for Vehicle Trajectory Prediction. 2018.
20. Bahari, M., I. Nejjar, and A. Alahi. Injecting Knowledge in Data-Driven Vehicle Trajectory Predictors. *Transportation Research Part C: Emerging Technologies*, 2021. https://doi.org/10.1016/j.trc.2021.103010.
21. Werling, M., and D. Liccardo. Automatic Collision Avoidance Using Model-Predictive Online Optimization. 2012.
22. Shen, C., H. Guo, F. Liu, and H. Chen. MPC-Based Path Tracking Controller Design for Autonomous Ground Vehicles. 2017.
23. Wang, H., Y. Huang, A. Khajepour, Y. Zhang, Y. Rasekhipour, and D. Cao. Crash Mitigation in Motion Planning for Autonomous Vehicles. *IEEE Transactions on Intelligent Transportation Systems*, 2019. https://doi.org/10.1109/TITS.2018.2873921.
24. Babu, M., R. R. Theerthala, A. K. Singh, B. P. Baladhurgesh, B. Gopalakrishnan, K. M. Krishna,

and S. Medasani. Model Predictive Control for Autonomous Driving Considering Actuator Dynamics. 2019.

25. Brockman, G., V. Cheung, L. Pettersson, J. Schneider, J. Schulman, J. Tang, and W. Zaremba. OpenAI Gym. 2016, pp. 1–4.

26. Nadimi, N., D. R. Ragland, and A. Mohammadian Amiri. An Evaluation of Time-to-Collision as a Surrogate Safety Measure and a Proposal of a New Method for Its Application in Safety Analysis. *Transportation Letters*, 2020. https://doi.org/10.1080/19427867.2019.1650430.

27. Xu, C., Z. Ding, C. Wang, and Z. Li. Statistical Analysis of the Patterns and Characteristics of Connected and Autonomous Vehicle Involved Crashes. *Journal of Safety Research*, 2019. https://doi.org/10.1016/j.jsr.2019.09.001.

28. Chen, S., Dong, J., Ha, P., Li, Y., Labi, S. Graph neural network and reinforcement learning for multi-agent cooperative control of connected autonomous vehicles. Computer-Aided Civil and Infrastructure Engineering, 2021. 36(7), 838-857.

29. Chen, S., Y. Leng, and S. Labi. A Deep Learning Algorithm for Simulating Autonomous Driving Considering Prior Knowledge and Temporal Information. *Computer-Aided Civil and Infrastructure Engineering*, 2019. https://doi.org/10.1111/mice.12495.

30. Dong, J., Chen, S., Y. Li, R. Du, A. Steinfeld, and S. Labi. Space-Weighted Information Fusion Using Deep Reinforcement Learning: The Context of Tactical Control of Lane-Changing Autonomous Vehicles and Connectivity Range Assessment. *Transportation Research Part C: Emerging Technologies*, 2021. https://doi.org/10.1016/j.trc.2021.103192.

31. Du, R., Chen, S., Li, Y., Ha, P., Dong, J., Labi, S., Anastasopoulos, PC. (2021). A cooperative crash avoidance framework for autonomous vehicle under collision-imminent situations in mixed traffic stream. IEEE Intelligent Transportation Systems Conference (ITSC), September 19-22, 2021. Indianapolis, IN, United States.

32. Dong, J., Chen, S., Zong, S., Chen, T., Labi, S. (2021). Image transformer for explainable autonomous driving system. IEEE Intelligent Transportation Systems Conference (ITSC), September 19-22, 2021. Indianapolis, IN, United States.

33. Du, R., Chen, S., Dong, J., Ha, P., Labi, S. (2021). GAQ-EBkSP: A DRL-based Urban Traffic Dynamic Rerouting Framework using Fog-Cloud Architecture. IEEE International Smart Cities Conference (ISC2), September 7-10, 2021 – Virtual Conference.

34. Dong, J., Chen, S., Li, Y., Du, R., Steinfeld, A., Labi, S. (2020). Spatio-weighted information fusion and DRL-based control for connected autonomous vehicles. IEEE Intelligent Transportation Systems Conference (ITSC), September 20 – 23, 2020. Rhodes, Greece.

35. Ha, P., Chen, S., Du, R., Dong, J., Li, Y., Labi, S. (2020). Vehicle connectivity and automation: a sibling relationship. Frontiers in Built Environment, 6, 199.